\documentclass{article}

\usepackage{microtype}
\usepackage{graphicx}
\usepackage{subfigure}
\usepackage{booktabs} % for professional tables
\usepackage{multirow} % Kai: Used for UCI-Table
\usepackage{multicol} % Kai: Used for UCI-Table

\usepackage{hyperref}

\usepackage[accepted]{icml2020}

\icmltitlerunning{Single Shot MC Dropout Approximation}

\begin{document}

\twocolumn[
\icmltitle{Single Shot MC Dropout Approximation}

% Submission and Formatting Instructions for
%           the ICML Workshop \\ on Uncertainty and Robustness in Deep Learning (ICML UDL 2020)



\begin{icmlauthorlist}
\icmlauthor{Kai Brach}{ebit}
\icmlauthor{Beate Sick}{zhaw_uzh}
\icmlauthor{Oliver D{\"u}rr}{htwg}
\end{icmlauthorlist}

\icmlaffiliation{ebit}{Elektrobit Automotive GmbH, Germany}
\icmlaffiliation{zhaw_uzh}{IDP, Zurich University of Applied Sciences, Switzerland, and EBPI, University of Zurich, Switzerland}
\icmlaffiliation{htwg}{IOS, Konstanz University of Applied Sciences, Germany}

\icmlcorrespondingauthor{Kai Brach}{kai.brach@elektrobit.com}
\icmlcorrespondingauthor{Oliver Dürr}{oliver.duerr@htwg-konstanz.de}
\icmlcorrespondingauthor{Beate Sick}{sick@zhaw.ch}

\icmlkeywords{Deep Neural Networks, Bayesian Neural Networks, Moment Propagation, Error Propagation, MC Dropout Approximation, Uncertainty}

\vskip 0.3in
]



\printAffiliationsAndNotice{\icmlEqualContribution} % otherwise use the standard text.

% Use input instead of include to prevent new page
\begin{abstract}

Deep neural networks (DNNs) are known for their high prediction performance, especially in perceptual tasks such as object recognition or autonomous driving. Still, DNNs are prone to yield unreliable predictions when encountering completely new situations  without indicating their uncertainty. Bayesian variants of DNNs (BDNNs), such as MC dropout BDNNs, do provide uncertainty measures. However, BDNNs are slow during test time because they rely on a sampling approach. Here we present a single shot MC dropout approximation that preserves the advantages of BDNNs without being slower than a DNN. Our approach is to analytically approximate for each layer in a fully connected network the expected value and the variance of the MC dropout signal.  We evaluate our approach on different benchmark datasets and a simulated toy example. We demonstrate that our single shot MC dropout approximation resembles the point estimate and the uncertainty estimate of the predictive distribution that is achieved with an MC approach, while being fast enough for real-time deployments of BDNNs.

\end{abstract}
\section{Introduction}
\label{sec:intro}
Over the last, decade deep neural networks (DNN) have arisen as the dominant technique for the analysis of perceptual data. Also in safety-critical applications like autonomous driving, where the vehicle must be able to understand its environment, DNNs have seen rapid progress in several tasks \cite{Grigorescu2019}.

However, classical DNNs have deficits in capturing the model uncertainty \cite{Kendall2017},\cite{Gal2016}. But when using DNN models in safety-critical applications, it is mandatory to provide an uncertainty measure that can be used to identify unreliable predictions \cite{Michelmore2018} \cite{Feng2018} \cite{Harakeh2019} \cite{Miller2018a} \cite{McAllister2017}.

For example, in the field of robotics \cite{Sunderhauf2018}, medical applications, or autonomous driving \cite{Bojarski2016}, where machines interact with humans, it is important to identify situations where a model prediction is unreliable and a human intervention is necessary. This can, for example, be situations which are completely different from all that occurred during training.

Employing Bayesian DNNs (BDNNs) \cite{MacKay1992} tackles the problem and allows to compute an uncertainty measure. However, state of the art BDNNs require sampling during deployment leading to computation times that are by the factor of MC runs larger than a classical DNNs. This work overcomes this drawback by providing a method that allows to approximate the expected value and variance of a BDNN's predictive distribution in a single run. It has therefore the same computation time as a classical DNN. We focus here on a special variant of BDNNs which is known as MC dropout \cite{Gal2016}. While our approximation method is applicable also to convolutional neural networks and classification settings, we focus in this work on regression through fully connected networks. 

Ensembling based models take an alternative approach to estimate uncertainties and have been successfully applied to DNNs \cite{lak2017,pearce2020}. But ensemble methods do also not allow to quantify the uncertainty in a single shot manner.

\section{Related Work}
\label{sec:related_work}
\subsection{MC Dropout Bayesian Neural Networks}
\label{subsec:related_work_mc}
BDNNs are probabilistic models that capture the uncertainty by means of probability distributions. Probabilistic DNNs, which are non-Bayesian, only define a distribution for the conditional outcome. 
In common probabilistic DNNs the output nodes are controlling the parameters of a conditional probability distribution (CPD) of the outcome. For regression type problems a common choice for the CPD is the normal distribution $N(\mu, \sigma^2)$, where the variance $\sigma^2$ quantifies the data uncertainty, known as aleatoric uncertainty. BDNNs define in addition distributions for the weights which translate in a distribution of the modeled parameters. In this manner  the model uncertainty is captured, which is known as epistemic uncertainty \cite{Der2009}. In case of MC dropout BDNNs each weight distribution is a Bernoulli distribution: the weight takes with the dropout probability $p^*$ the value zero and with probability $1-p^*$ the value $w$. All weights starting from the same neuron are set to zero simultaneously. The dropout probability $p^*$ is usually treated as a fixed hyperparameter and the weight-value $w$ is tuned during the training. 

In contrast to standard dropout \cite{Srivastava2014}, the weights in MC dropout are not frozen and rescaled after training, but  the dropout procedure is also done during test time. It can be shown that MC dropout is an approximation to a BDNN \cite{Gal2016}. MC dropout BDNNs were successfully used in many applications and have proven to yield improved prediction performance and allow to define uncertainty measures to identify individual unreliable predictions \cite{Gal2016}, \cite{Ryu2019}, \cite{Duerr2018}, \cite{Kwon2020}. To employ a trained Bayesian DNN in practice one performs several runs of predictions. 
In each run, weights are sampled from the weight distributions leading to a certain constellation of weight values that are used to compute the parameters of a CPD. To determine the outcome distribution of a BDNN, we draw samples from the CPDs that resulted from different MC runs. In this way, the  outcome distribution incorporates the epistemic and aleatoric uncertainty. A drawback of a MC dropout BDNN compared to its classical DNN variant is the increased computing time. The sampling procedure leads to a computing time that is prohibitive for many real-time applications like autonomous driving.

\subsection{Moment Propagation}
\label{subsec:related_work_ep}
Our method relies on statistical moment propagation (MP). More specifically, we propagate the expectation and the variance, of our signal distribution through the different layers of a neural network. The variance of the signal arises due to the dropout process. Quantifying the variance after a transformation is also done in error propagation (EP). EP quantifies how an uncertainty of an input which is transformed by a function (i.e. a measurement error) transfers to an uncertainty of the output of this function. In case of a continuous output it is common to characterize the uncertainty by the variance. This approach is also used in statistics as the delta method \cite{Dorfman1938}. In MP we approximate the layer-wise transformations of the variance and the expected value. A similar approach has also been used for neural networks before \cite{Frey1998,adachi2019}, and used to detect adversarial examples in \cite{Jin2015} and \cite{Gast2018}. 

But, due to our best knowledge, our approach is the first method that provides a single shot approximation to the expected value and the variance of the predictive distribution resulting from a MC dropout NN.

\section{Methods}
\label{sec:methods}
\begin{figure}[h]
    \centering
    \includegraphics[width=0.45\textwidth]
    {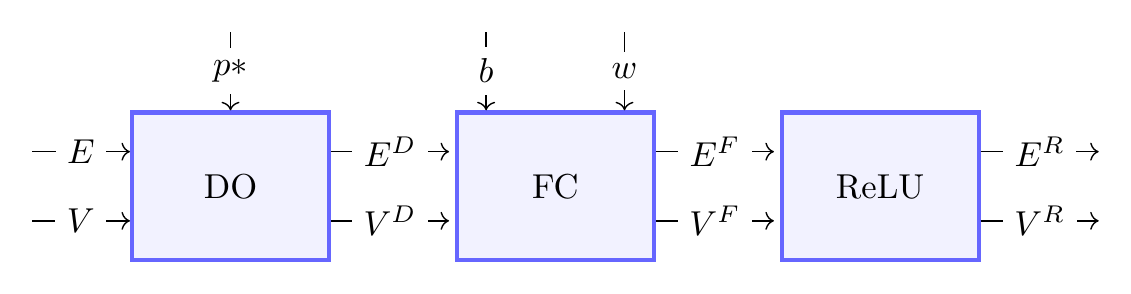}
    \caption{Overview of the proposed method. The expectation E and V flow through different layers of the network in a single forward pass. Shown is an example configuration in which Dropout (DO) is followed by Dense (FC) and a ReLU activation. More complex networks can be build by different arrangements of the individual blocks. }
    \label{fig:over}
\end{figure}

The goal of our method\footnote{\href{https://github.com/kaibrach/Moment-Propagation}{https://github.com/kaibrach/Moment-Propagation}} is to approximate the expected value E and the variance V of the predicted output which is obtained by the above described MC dropout method. When propagating an observation through a MC dropout network, we get each layer with $p$ nodes an activation signal with an expected value $E$ (of dimension $p$) and a variance given by a variance-covariance matrix $V$ (of dimension $p \times p$). We neglect the effect of correlations between different activations, which are small anyway in deeper layers due to the decorrelation effect of the dropout. Hence, we only consider diagonal terms in the correlation matrix. In the following, we describe for each layer-type in a fully connected network how the expected value E and its variance V is propagated. As layer-type we consider dropout, dense, and ReLU activation layer. Figure \ref{fig:over} provides an overview of the layer-wise abstraction.

\subsection{Dropout Layer}
We start our discussion, with the effect of MC dropout. Let $E_i$ be the expectation at the i’th node of the input layer and $V_i$ the variance at the i’th node. In a dropout layer the random value of a node $i$ is multiplied independently with a Bernoulli variable $Y \sim \tt{Bern(p^*)}$ that is either zero or one. The expectation $E_i^D$ of the i'th node after dropout is then given by:

\begin{equation}
E^D_i = E_i(1-p^*)
\label{eq:e_do}    
\end{equation}

For computing the variance $V_i^D$ of the i'th node after dropout, we use the fact that the variance $V(X \cdot Y)$ of the product of two independent random variables $X$ and $Y$, is given by \cite{Goodman1960}:

\begin{equation}
    V(X \cdot Y) = V(X) V(Y) + V(X) E^2(Y) + E^2(X) V(Y)
\end{equation}

With $V(Y)=p^*(1-p^*)$, we get:
\begin{equation}
V_i^D=V_i \cdot p^*(1-p^*)+V_i (1-p^*)^2+E^2_i \cdot p^*(1-p^*)
\label{eq:v_do}    
\end{equation}

Dropout is the only layer in our approach where uncertainty is created. I.e. even if the input has $V_i = 0$  the output of the dropout layer has $V_i^D > 0$ for $p^* \ne 0$.

\subsection{Dense Layer}
For the dense layer with $p$ input and $q$ output nodes, we compute the value of the i'th output node  as $\sum_j^p w_{ji} x_j + b_i$, where $x_j, j=1 \dots p$ are the values of the input nodes. Using the linearity of the expectation, we get the expectation $E^F_i$ of the i'th output node from the expectations, $E^F_j, j=1 \dots p$, of the input nodes:
\begin{equation}
E^F_i =\sum_{j=1}^p w_{ji} E_j + b_i
\label{eq:e_fc}    
\end{equation}
To calculate the change of the variance, we use the fact that the variance under a linear transformation behaves like $V( w_{ji} \cdot x_j + b)= w_{ji}^2 V(x_j)$. Further, we assume independence of the j different summands, yielding:  
\begin{equation}
V_i^F = \sum_{j=1}^p w^2_{ji} V_j
\label{eq:v_fc}    
\end{equation}

\subsection{ReLU Activation Layer}
To calculate the expectation $E^{R}_i$  and variance $V_i^R$ of the i'th node after a ReLU, as a function of the $E_i$ and $V_i$ of this node before the ReLU, we need to make a distributional assumption.  We assume that the input is Gaussian distributed, with $\phi(x)=N(x;E_i,V_i)$ the PDF, and $\Phi(x)$ the corresponding CDF, we get (see \cite{Frey1998} for a derivation) for the expectation and variance of the output:

\begin{equation}
E^R_i = E_i \cdot \Phi\left(\frac{E_i}{\sqrt{V_i}}\right) + \sqrt{V_i} \cdot \phi\left(\frac{E_i}{\sqrt{V_i}}\right) 
\label{eq:e_relu}    
\end{equation}

\begin{equation}
% V^R_i = (\mu^2 + \sigma^2) \Phi\left(\frac{\mu}{\sigma}\right) + \mu \sigma \cdot \phi\left(\frac{\mu}{\sigma}\right) - (E^R_i)^2 
V^R_i = (E_i^2 + V_i) \cdot \Phi\left(\frac{E_i}{\sqrt{V_i}}\right) + E_i \sqrt{V_i} \cdot \phi\left(\frac{E_i}{\sqrt{V_i}}\right) - {E^R_i}^2
\label{eq:v_relu}    
\end{equation}

\section{Results}
\label{sec:results}

\subsection{Toy Dataset}
\begin{figure}[h]
    \centering
    \includegraphics[width=0.45\textwidth]
    {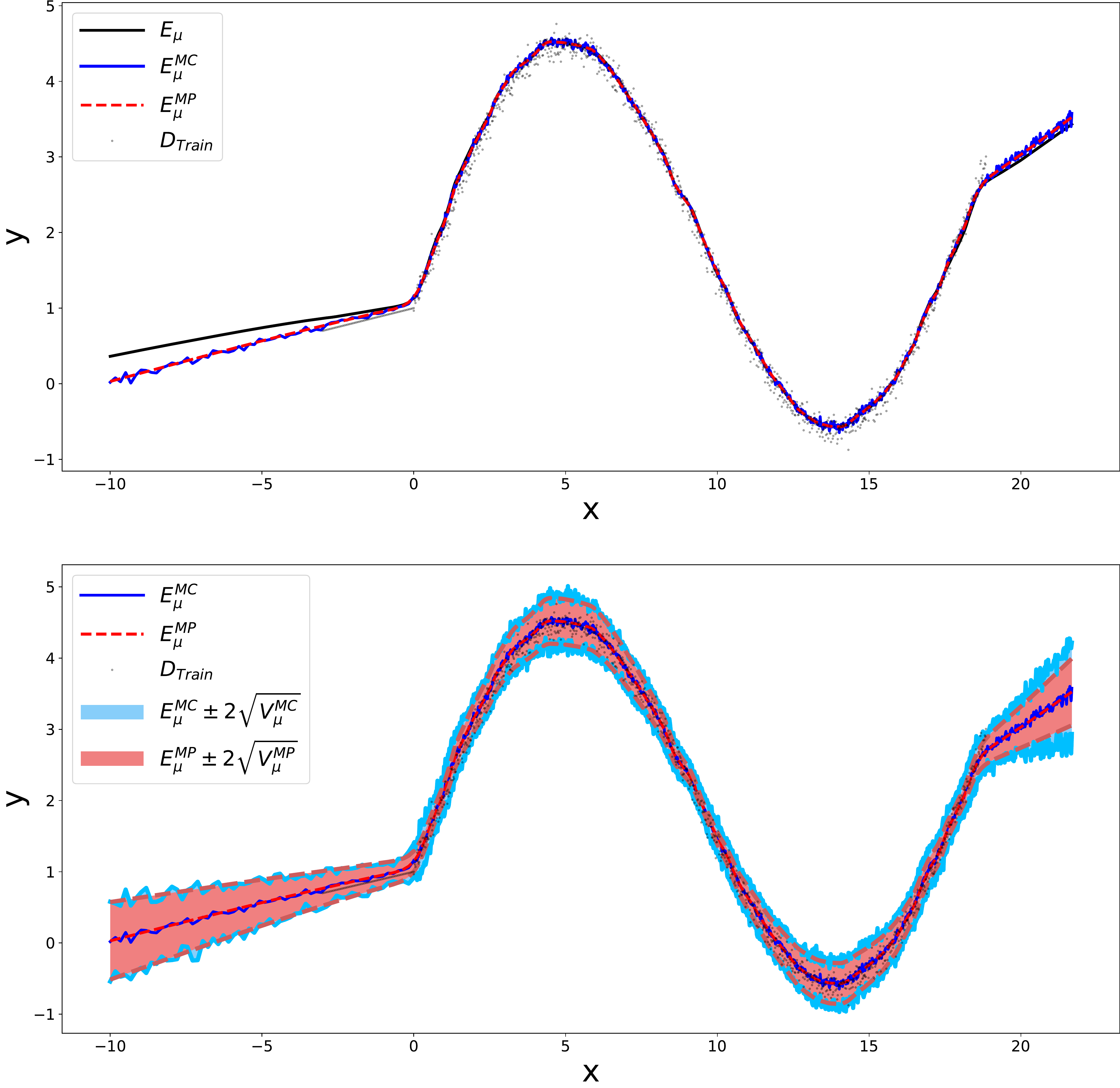}
    \caption{Comparison of the MP and MC dropout results of a BDNN and the results of a DNN. The NNs were fitted on train data that were available in the range of -3 to 19. In the upper panel the estimated expectations of the MC BDNN, the MP BDNN, and the DNN are compared. In the lower panel the predicted spread of $\mu(t)$ is shown for the MC and MP method.}
    \label{fig:toy}
\end{figure}

We first apply our approach to a one dimensional regression toy dataset, with only one input feature. We use a fully connected NN with three layers each with 256 nodes, ReLU activations and dropout after the dense layers. We have a single node in the output layer which is interpreted as the expected value $\mu$ of the conditional outcome distribution $p(y|x)$. We train the network using the MSE loss and apply dropout with $p^*=0.3$. From the MC dropout BDNN, we get at each x-position $T=30$ MC samples $\mu_t(x)$ from which we can estimate the expectation $E_\mu$ by the average value and $V_\mu$ by the variance of $\mu_t(x)$. For comparison, we use our MP approach to also approximate the expected value $E_\mu$ and the variance $V_\mu$ of $\mu$ at each $x$-position (see upper panel of  \ref{fig:toy}). We also included the deterministic output $\mu(x)$ of the DNN in which dropout has only been used only during training. All three approaches yield nearly identical results, within the range of the training data. We attribute this to the fact, that we have plenty of training data and so the epistemic uncertainty is neglectable. In the lower panel of figure \ref{fig:toy} a comparison of the uncertainty of $\mu(x)$ is shown by displaying an interval given by the expected value of $\mu(x)$ plus-minus two times the standard deviation of $\mu(x)$. Here the width of the resulting intervals of a BDNN via the MP approach and the MC dropout are comparable (the DNN has no spread). This indicates the usefulness of this approach for epistemic uncertainty estimation.  

% UCI Dataset Evaluation
\begin{table*}[t]
	\caption{Comparison of the average prediction performance in test RMSE (Root-Mean-Square Error), test NLL (Negative Log-Likelihood) and test RT (Runtime) including $\pm$ standard error on UCI regression benchmark datasets between MC and MP. \textit{N} and \textit{Q} correspond to the dataset size and the input dimension. For all test measures, smaller means better.}
	\label{tbl:uci_benchmark}
	\vskip 0.15in
	\begin{center}
		\begin{small}
			\begin{sc}
				\resizebox{\linewidth}{!}{\begin{tabular}{lllcccccc}
						\toprule
						\multirow{2}{*}{Dataset} &
						\multirow{2}{*}{\textit{N}} &
						\multirow{2}{*}{\textit{Q}} &
						\multicolumn{2}{c}{Test RMSE} &
						\multicolumn{2}{c}{Test NLL} &
						\multicolumn{2}{c}{Test RT [s]} \\
						& & & {MC} & {MP} & {MC} & {MP} & {MC} & {MP} \\
						\midrule
	                    Boston		&506	&13 &3.14 $\pm{0.20}$	&3.10 $\pm{0.20}$	&2.57 $\pm{0.07}$	&2.56 $\pm{0.08}$	&2.51 $\pm{0.03}$	&0.04 $\pm{0.00}$\\
						Concrete	&1,030	&8 	&5.46 $\pm{0.12}$	&5.40 $\pm{0.12}$	&3.12 $\pm{0.02}$	&3.13 $\pm{0.03}$	&3.37 $\pm{0.04}$	&0.04 $\pm{0.00}$\\	
						Energy		&768	&8 	&1.65 $\pm{0.05}$	&1.61 $\pm{0.05}$	&1.95 $\pm{0.04}$	&2.01 $\pm{0.04}$	&2.84 $\pm{0.03}$	&0.04 $\pm{0.00}$\\	
						Kin8nm		&8,192	&8	&0.08 $\pm{0.00}$	&0.08 $\pm{0.00}$	&-1.10 $\pm{0.01}$	&-1.11 $\pm{0.01}$	&7.37 $\pm{0.06}$	&0.04 $\pm{0.00}$\\	
						Naval		&11,934	&16 &0.00 $\pm{0.00}$	&0.00 $\pm{0.00}$	&-4.36 $\pm{0.01}$	&-3.60 $\pm{0.01}$	&9.69 $\pm{0.11}$	&0.04 $\pm{0.00}$\\
						Power		&9,568	&4 	&4.05 $\pm{0.04}$	&4.04 $\pm{0.04}$	&2.82 $\pm{0.01}$	&2.84 $\pm{0.01}$	&6.85 $\pm{0.07}$	&0.04 $\pm{0.00}$\\		
						Protein		&45,730	&9  &4.42 $\pm{0.03}$	&4.41 $\pm{0.02}$	&2.90 $\pm{0.00}$	&2.91 $\pm{0.00}$	&31.38 $\pm{0.09}$	&0.05 $\pm{0.00}$\\
						Wine		&1,599	&11 &0.63 $\pm{0.01}$	&0.63 $\pm{0.01}$	&0.95 $\pm{0.01}$	&0.95 $\pm{0.01}$	&4.78 $\pm{0.01}$	&0.04 $\pm{0.00}$\\		
						Yacht		&308	&6 	&2.93 $\pm{0.22}$	&2.91 $\pm{0.26}$	&2.35 $\pm{0.07}$	&2.11 $\pm{0.07}$	&2.01 $\pm{0.01}$	&0.04 $\pm{0.00}$\\
						\bottomrule
				\end{tabular}}
			\end{sc}
		\end{small}
	\end{center}
	\vskip -0.1in
\end{table*}

\subsection{UCI-Datasets}
To benchmark our method, we redo the analysis of \cite{Gal2016} for the UCI regression benchmark dataset. We use the same NN model as Gal and Ghahramani, which is a fully connected neural network including one hidden layer with ReLU activation in which the CPD $p(y|x)$ over $T=10,000$ MC runs is given by sampling from the normal PDF:

\begin{equation}
p(y|x) = \frac{1}{T} \sum_t N(y ; \mu_t(x), \tau^{-1})
\label{eq:gal_cpd}    
\end{equation}

Again $\mu_t(x)$ is the single output of the BDNN for the t'th MC run. To derive a predictive distribution Gal assumes in each run a Gaussian distribution, centered at $\mu$ and a precision $\tau$, corresponding to the reciprocal of the variance. The parameter $\mu$ is received from the NN and $\tau$ is treated as as a hyperparameter. For the MP model, the MC sampling (Eq. \ref{eq:gal_cpd}) is replaced by integration: 

\begin{eqnarray}
p(y|x) &=& \int N(y ; \mu', \tau^{-1}) N(\mu'; E\textsuperscript{MP}, V\textsuperscript{MP}) \; d\mu'\nonumber\\ 
&=& N(y;E\textsuperscript{MP}, V\textsuperscript{MP} + \tau^{-1})
\label{eq:mp_cpd}    
\end{eqnarray}

We used the same protocol as \cite{Gal2016} which can be found at \href{https://github.com/yaringal/DropoutUncertaintyExps}{https://github.com/yaringal/DropoutUncertaintyExps}. Accordingly, we train the network for 10$\times$ the epochs provided in the individual dataset configuration. As described in \cite{Gal2016} an excessive  grid search over the dropout rate $p^*=0.005,0.01,0.05,0.1$ and different values of the precision $\tau$ is done. The hyperparameters minimizing the validation NLL are chosen and applied on the testset. 

We report in table \ref{tbl:uci_benchmark} the test performance (RMSE and NLL) achieved via MC BDNN using the optimal hyperparameters  for the different UCI datasets. We also report the test RMSE and the NLL achieved with our MP method. Allover, the MC and MP approaches  produces similar results. However, as shown in the last column in the table the MP method is much faster, having only to perform one forward pass instead of $T=10,000$ forward passes. 
\section{Discussion}
\label{sec:discussion}
With our MP approach we have introduced an approximation to MC dropout which requires no sampling but instead propagates the expectation and the variance of the signal through the network. This results in a time saving by a factor that approximately corresponds to the number of MC runs (in our benchmark experiment 10,000). We have shown that our fast MP approach approximates precisely the expectation and variance of the prediction distribution achieved by MC dropout. Also the achieved prediction performance in terms of RMSE and NLL do not show significant differences when using MC dropout or our MP approach. Hence, our presented MP approach opens the door to include uncertainty information in real-time applications.

We are currently working on extending the approach to different architectures such as convolutional neural networks.We are also investigating how to make use of the uncertainty information to detect novel classes in classification settings.

\section{Acknowledgements}
\label{sec:acknowledgements}

We are very grateful to Elektrobit Automotive GmbH for supporting this research work. Further, part of the work has been founded by the Federal	Ministry of	Education	and	Research	of	Germany	(BMBF) in the project DeepDoubt (grant no. 01IS19083A).


% kai: Enable later
\bibliography{error_propagation_paper}
\bibliographystyle{icml2020}

%\newpage
%\appendix
%\input{A1}

\end{document}